\documentclass[runningheads,a4paper]{llncs}

\usepackage{amssymb}
\usepackage{amsmath}
\usepackage{graphicx}
\usepackage{adjustbox}

\usepackage{url}
\urldef{\mailsa}\path|l.peel@cs.ucl.ac.uk|    
\newcommand{\keywords}[1]{\par\addvspace\baselineskip
\noindent\keywordname\enspace\ignorespaces#1}

\begin{document}

\mainmatter  

\title{Supervised Blockmodelling}

\titlerunning{Supervised Blockmodelling}

%
%
\author{Leto Peel}
\authorrunning{Supervised Blockmodelling}

\institute{Department of Computer Science, University College London,\\
London, UK\\
\mailsa\\
}
%
%

\maketitle

\begin{abstract}
Collective classification models attempt to improve classification performance by taking into account the class labels of related instances.  However, they tend not to learn patterns of interactions between classes and/or make the assumption that instances of the same class link to each other (assortativity assumption).  Blockmodels provide a solution to these issues, being capable of modelling assortative and disassortative interactions, and learning the pattern of interactions in the form of a summary network.  The Supervised Blockmodel provides good classification performance using link structure alone, whilst simultaneously providing an interpretable summary of network interactions to allow a better understanding of the data.  This work explores three variants of supervised blockmodels of varying complexity and tests them on four structurally different real world networks.
\end{abstract}

\section{Introduction}
Probabilistic classification algorithms have long focused on the problem of predicting unknown labels of data instances according to their attributes by leveraging the conditional distributions of a supplied training set.  These algorithms traditionally made an assumption that the data was independent and identically distributed, however many modern datasets break this assumption. As a result, research has shifted to examining how these relations or links can be exploited to improve classification performance.  Collective classification is one approach which attempts this but often assumes that instances of a given class tend to link others of the same class, i.e that the class instances are assortative. 

Recently, stochastic blockmodelling has been applied in a classification context to overcome the need for assortativity assumptions \cite{Peel:Fusion11,Moore:2011}.  In addition, the stochastic blockmodel can be used to understand the pattern of interactions between class instances by the way of a summary network of \textit{role} interactions.  This work describes three classification models of varying complexity based on the stochastic blockmodel along with efficient inference updates using collapsed variational Bayes. Their relative performance is investigated in various within-network classification cases.  Finally, an example is given of the analysis that can be conducted on the resulting model to better understand both the structure of the data and the classification decision.  

The main contribution of this work is the comparison between the models and the introduction of a model of intermediate complexity (section \ref{sec:supsingmodel}).  A minor contribution is the new update equations (section \ref{sec:colvarinf}) based on collapsed variational inference \cite{TehNewWel2007} which avoids the long running time and convergence diagnosis of the Gibbs sampling in \cite{Moore:2011} and the parameter updates and expensive Digamma function evaluations of the variational inference in \cite{Peel:Fusion11}.

\section{Background}
Blockmodels have been used for social and psychometric analysis for decades \cite{LfWh71,Holland1976,wassermanFaust}.  The name refers to the ``blocks'' of zero and non-zero elements that occur in the adjacency matrix when the rows and columns are reordered such that nodes with similar interaction patterns are adjacent. The clusters of nodes which make up these block patterns are known as network roles.  Nodes belonging to the same role are equivalent to each other with respect to their probability of linking to nodes of other roles in the network. The pattern of interactions between roles provides a summary of the interactions of the network.  The original  blockmodels were \textit{a-priori} blockmodels where the assignment of nodes to roles was predetermined, usually according to the attributes of the nodes.
Bayesian formulations of blockmodelling, known as stochastic blockmodelling \cite{nowicki2001eap}, were developed to create blockmodels by automatically inferring nodes roles according to the posterior distribution given the observed network links.  The blockmodelling paradigm then comes full circle with the supervised blockmodel as the roles inferred from the network structure are then used to predict the attributes of the network in a given classification problem.   

Stochastic blockmodels are usually used in an unsupervised context but can easily be transferred to the supervised setting by simply instantiating the roles of the nodes in the training set and inferring the remainder of the network as in \cite{Moore:2011}.  In this case no extra variables are required as the roles and classes are equivalent and the inference procedure remains the same.  This type of model, however, assumes that the classes are homogeneous in their linkage patterns and that all nodes of a particular class behave in the same way.  To address this, an extension to the standard blockmodelling approach can be made based on supervised Latent Dirichlet Allocation (sLDA) \cite{blei:sLDA} which, as the name suggests, is a supervised extension of the topic modelling approach LDA \cite{Blei:LDA}. Latent Dirichlet Allocation is a method for clustering a corpus of documents into topics.  The sLDA model extends the LDA approach to identify topics in documents which not only best describes the document structures but also to predict a known response variable (i.e. a classification or regression target) associated with each document.  Similarly, a supervised blockmodel can be derived which identifies roles which both summarise the network structure and predict the class labels of nodes.  By making a distinction between the roles and the classes it is possible to model heterogeneous linkage patterns within classes.  Two such models are presented here: one which assigns a single role to each node, and one which allows nodes to have multiple role memberships.


\section{Supervised Blockmodels}

This section describes the three variants of supervised blockmodels examined in this paper.  

\subsection{Standard Stochastic Blockmodel}
A standard Stochastic Blockmodel (SBM) assumes the following generative process:

\begin{enumerate}
	\item For a given network draw a distribution over the $K$ roles in the network $\theta \sim Dirichlet(\alpha)$
	\item For each of the $K \times K$ possible role interactions:
	\begin{enumerate}
		\item Draw a probability of interacting $\pi_{k_1,k_2} \sim Beta(\beta_1,\beta_2)$
	\end{enumerate}
	\item For each node in the network, $v \in \{1,..,N\}$: 
	\begin{enumerate}
		\item Draw a role $z_v \sim Categorical(\theta)$
	\end{enumerate}
	\item For each of the $N \times N$ possible sender-receiver directed interactions, $s,r$:
	\begin{enumerate}
		\item Draw a binary value to indicate the presence or absence of a link $e_{s,r} \sim Bernoulli(\pi_{z_s,z_r})$
	\end{enumerate} 
\end{enumerate}

The application of a standard Stochastic Blockmodel was demonstrated in \cite{Moore:2011} where the roles and classes are considered as the same thing.  

\subsection{Supervised Single Membership Blockmodel}
\label{sec:supsingmodel}

The Supervised Single Membership Blockmodel (SSMB) is very similar to the SBM but introduces a separate class variable to allow for heterogeneity within classes, i.e. each class may have more than one role.

\begin{enumerate}
	\item For a given network draw a distribution over the $K$ roles in the network $\theta \sim Dirichlet(\alpha)$
	\item For each of the $K \times K$ possible role interactions:
	\begin{enumerate}
		\item Draw a probability of interacting $\pi_{k_1,k_2} \sim Beta(\beta_1,\beta_2)$
	\end{enumerate}
	\item For each role in the network, $k \in \{1,...,K\}$:
	\begin{enumerate}
		\item Draw a distribution over classes $\mu_k \sim Dirichlet(\eta)$
	\end{enumerate}
	\item For each node in the network, $v \in \{1,..,N\}$: 
	\begin{enumerate}
		\item Draw a role $z_v \sim Categorical(\theta)$
		\item Draw a class label $ y_v \sim Categorical(\mu_{z_v})$
	\end{enumerate}
	\item For each of the $N \times N$ possible sender-receiver directed interactions, $s,r$:
	\begin{enumerate}
		\item Draw a binary value to indicate the presence or absence of a link $e_{s,r} \sim Bernoulli(\pi_{z_s,z_r})$
	\end{enumerate} 
\end{enumerate}

\subsection{Supervised Mixed Membership Blockmodel}
The Supervised Mixed Membership Blockmodel (previously presented in \cite{Peel:Fusion11}) extends the unsupervised mixed membership blockmodels from the literature \cite{Airoldi:MMSB,Sinkkonen:Component,DuBois:2010}. The Supervised Mixed Membership Blockmodel (SMMB) assumes the following generative process:
\begin{enumerate}
	\item For a given network draw a distribution over the possible $K^2$ network role interactions $\pi \sim Dirichlet(\alpha)$
	\item For each role $k \in \{1,2,...,K\}$:
	\begin{enumerate}
		\item Draw a distribution over nodes $\phi \sim$ $ Dirichlet(\beta) $
	\end{enumerate}
	\item For each interaction $i \in \{1,2,...,I\}$:
	\begin{enumerate}
		\item Draw a role interaction pair $z_i = (z_s,z_r)_i$, $z_i \sim Categorical(\pi)$
		\item Draw a sender node $s_i \sim Categorical(\phi_{z_s})$
		\item Draw a receiver node $r_i \sim Categorical(\phi_{z_r})$
	\end{enumerate}
	\item For each node $v \in \{1,2,...,N\} $:
	\begin{enumerate}
		\item Draw a class label $y_v \sim Softmax(\eta,\bar{z}_v)$ 
			\end{enumerate}
\end{enumerate}
		where $\bar{z}_v = \frac{1}{n_v}\sum_i{z_{s_i}\delta_{s_i,v} + z_{r_i}\delta_{r_i,v}}$ and $z_{s_i}$ and $z_{r_i}$ are the indicator vectors of length $K$ describing the network role of the sender and receiver nodes  in interaction $i$,  $\delta_{\cdot,\cdot}$ is the Kronecker delta.   $\bar{z}_v$ therefore represents the empirical behavior class frequencies for node $v$.  The softmax function provides the following distribution:
		
			$\quad p(y_v|\eta,\bar{z}_v) = \exp(\eta_{y_v}^T\bar{z}_v)/\sum_c{\exp(\eta_c^T\bar{z}_v)}$
\hspace{5mm}

\section{Collapsed Variational Inference}
\label{sec:colvarinf}
Inference of the network roles, $\mathbf{z}$, can be efficiently computed using variational inference.  Variational inference has the advantage over sampling methods due to convergence that is faster and easier to diagnose.  Previous work has shown that the performance differences between inference methods can be minimal given appropriate hyperparameter settings \cite{AsuWelSmy2009a}.  

Variational Bayes \cite{attias00avariational} introduces an approximate variational posterior distribution, $q$, over the latent variables (roles) and model parameters $(\mathbf{\pi}, \mathbf{\phi}, \mathbf{\theta})$.  Usually this is a fully factorised distribution known as a mean-field approximation which provides a more tractable lower bound on the log evidence.  
\begin{equation}
	\log p(\mathbf{x}|\Theta) \geq \mathcal{F}(q,\Theta) = E_q[\log p(\mathbf{z,x}|\Theta)] - E_q[q(\mathbf{z})]. 
	\label{freeE}
\end{equation}
By taking advantage of the conjugacy of the Dirichlet-Categorical and Beta-Bernoulli distributions, the model parameters can be integrated out exactly.  This treatment yields the collapsed variational posterior, parameterised by the variational parameter $\lambda$:
\begin{equation}
	q(z) = \prod_i{q(z_i|\lambda_i)},
\label{eq:collpost}
\end{equation}   
which provides a tighter bound on the evidence \cite{TehNewWel2007}.  However, exact implementation of collapsed variational Bayes is computationally too expensive and therefore in practice a first order Taylor expansion is used to approximate the update equations.
Further information on collapsed variational inference along with the first order approximation implemented here is given in \cite{Sung:2008:LVB,AsuWelSmy2009a}.  The following sections detail the update equations for the 3 models.

\subsection{Standard Stochastic Blockmodel}
Inference in the standard Stochastic Blockmodel consists of sequentially updating the variational posterior distribution over role assignments for each node according to:
\small
\begin{align}
	\lambda_{v,k}& \propto   
	(n_k + \alpha)\notag\\
	\times & \frac{ \prod_{i=1}^{f_{v,k}+g_{vk}}{\left(d_{k,k}^{\neg v} + \beta_1 + i\right)} \prod_{i=1}^{2n_{k}^{\neg v}-f_{v,k}-g_{v,k}+1}{\left(\left(n_{k}^{\neg v}\right)^2 - d_{k,k}^{\neg v} + \beta_2 + i\right)} }{\prod_{i=1}^{2n_{k}^{\neg v}+1}{\left(\left(n_{k}^{\neg v}\right)^2 + \beta_1 + \beta_2 + i\right)}}\notag\\
	\times &\prod_{k_2}{\left(\frac{ \prod_{i=1}^{f_{v,k}}{\left(d_{k,k_2}^{\neg v} + \beta_1 + i\right)} \prod_{i=1}^{n_{k_2}^{\neg v}-f_{v,k}}{\left(n_k^{\neg v}n_{k_2}^{\neg v} - d_{k,k_2}^{\neg v} + \beta_2 + i\right)} }{\prod_{i=1}^{n_{k_2}^{\neg v}}{\left(n_k^{\neg v}n_{k_2}^{\neg v} + \beta_1 + \beta_2 + i\right)}}\right)^{(1-\delta_{k,k_2})}} \notag \\
	\times &\prod_{k_1}{\left(\frac{ \prod_{i=1}^{g_{v,k}}{\left(d_{k_1,k}^{\neg v} + \beta_1 + i\right)} \prod_{i=1}^{n_{k_1}^{\neg v}-g_{v,k}}{\left(n_{k_1}^{\neg v}n_{k}^{\neg v} - d_{k_1,k}^{\neg v} + \beta_2 + i\right)} }{\prod_{i=1}^{n_{k_1}^{\neg v}}{\left(n_{k_1}^{\neg v}n_{k}^{\neg v} + \beta_1 + \beta_2 + i\right)}}\right)^{(1-\delta_{k_1,k})}} ,
\label{eq:SSBupdate}
\end{align}
\normalsize
where $d_{k_1,k_2}$ is the count of links from role $k_1$ to role $k_2$, $f_{v,k}$ is the number of times node $v$ sender to a node of role $k$ and similarly $g_{v,k}$ is the number of time $v$ is a receiver in an interaction with a node of role $k$.  The totals for each role are given by $n_k$.  Collapsed variational inference involves removing the counts of the current node which is denoted by $\neg v$.  
The nodes in the network used for training have their roles initialised to reflect their class (i.e. role=class) and the inference is carried out on the unlabelled nodes only.

\subsection{Supervised Single Membership Blockmodel}
The update procedure for the SSMB is almost identical to the SBM but incorporates information about the class-role co-occurrence counts $m_{c,k}$.  
\small
\begin{align}
\lambda_{v,k}& \propto (n_k + \alpha)\frac{\left(m_{y_v,k}^{\neg v}+\eta\right)}{\left(m_{\cdot,k}^{\neg v}+C\eta\right)}\notag\\
	\times & \frac{ \prod_{i=1}^{f_{v,k}+g_{v,k}}{\left(d_{k,k}^{\neg v} + \beta_1 + i\right)} \prod_{i=1}^{2n_{k}^{\neg v}-f_{v,k}-g_{v,k}+1}{\left(\left(n_{k}^{\neg v}\right)^2 - d_{k,k}^{\neg v} + \beta_2 + i\right)} }{\prod_{i=1}^{2n_{k}^{\neg v}+1}{\left(\left(n_{k}^{\neg v}\right)^2 + \beta_1 + \beta_2 + i\right)}}\notag\\
	\times& \prod_{k_2}{\left(\frac{ \prod_{i=1}^{f_{v,k}}{\left(d_{k,k_2}^{\neg v} + \beta_1 + i\right)} \prod_{i=1}^{n_{k_2}^{\neg v}-f_{v,k}}{\left(n_k^{\neg v}n_{k_2}^{\neg v} - d_{k,k_2}^{\neg v} + \beta_2 + i\right)} }{\prod_{i=1}^{n_{k_2}^{\neg v}}{\left(n_k^{\neg v}n_{k_2}^{\neg v} + \beta_1 + \beta_2 + i\right)}}\right)^{(1-\delta_{k,k_2})}} \notag \\
	\times &\prod_{k_1}{\left(\frac{ \prod_{i=1}^{g_{v,k}}{\left(d_{k_1,k}^{\neg v} + \beta_1 + i\right)} \prod_{i=1}^{n_{k_1}^{\neg v}-g_{v,k}}{\left(n_{k_1}^{\neg v}n_{k}^{\neg v} - d_{k_1,k}^{\neg v} + \beta_2 + i\right)} }{\prod_{i=1}^{n_{k_1}^{\neg v}}{\left(n_{k_1}^{\neg v}n_{k}^{\neg v} + \beta_1 + \beta_2 + i\right)}}\right)^{(1-\delta_{k_1,k})}} 
\label{eq:SMSBupdate}
\end{align}
\normalsize
In this model the roles are initialised randomly and the inference procedure sequentially updates each node in the network until convergence. Note that as opposed to the SBM model inference occurs over the training nodes too.  

Prediction of the unknown labels requires calculation of the parameter $\mathbf{\mu}$ according to \eqref{eq:mu}:
\small
\begin{equation}
	\widehat{\mu}_{c,k} = \frac{(m_{c,k} + \eta)}{(m_{\cdot,k} + C\eta)}
\label{eq:mu}
\end{equation}
Prediction of node class labels is done according to \eqref{eq:predSMSB}
\begin{equation}
	y_v^* = \arg \max_{y \in \{1,...,C\}} E_q \left[ \mu_y^T \bar{z_v} \right] = \arg \max_{y \in \{1,...,C\}} \mu_y^T \bar{\lambda}_v.
\label{eq:predSMSB}
\end{equation}
\normalsize
\subsection{Supervised Mixed Membership Blockmodel}
For the SMMB, role pairs (sender role and receiver role) are assigned to each interaction rather than assigning a single role to a node.  Inference is therefore conducted by sequentially updating the variational posterior for each network interaction according to:
\small
\begin{align}
	\lambda_{i,k_1,k_2} \propto &  \left( d_{k_1,k_2}^{\neg i} + \alpha_{k_1,k_2} \right) \frac{(f^{\neg i}_{s_i,k_1} + \beta)(g^{\neg i}_{r_i,k_2} + \beta)}{(f^{\neg i}_{\cdot, k_1} + N\beta)(g^{\neg i}_{\cdot,k_2} + N\beta + \delta_{k_1,k_2})}   \notag\\
  &  \times \exp \left( \frac{\eta_{y_{s_i},k_1}}{n_{s_i}}-\frac{h_{i,s_i,k_1}}{h_{i,s_i}^T\lambda_{s_i}^{old}} +   \frac{\eta_{y_{r_i},k_2}}{n_{r_i}}-\frac{h_{i,r_i,k_2}}{h_{i,r_i}^T\lambda_{r_i}^{old}} \right) ,
\end{align}
\normalsize
where $f_{v,k}$ ($g_{v,k}$) are the number of times node $v$ is a sender (receiver) as role $k$, $n_{v}$ is the number of times node $v$ is involved in an interaction and $\lambda_{v}$ is a $K$-length vector representing the marginal probability of sender or receiver positions, i.e.:
\small
\begin{align}
 \lambda_{s_i} = \left[\sum_{k_2}{\lambda_{i,1,k_2}}, \sum_{k_2}{\lambda_{i,2,k_2}}, \cdots, \sum_{k_2}{\lambda_{i,K,k_2}}\right]^T, \notag\\
 \lambda_{r_i} = \left[\sum_{k_1}{\lambda_{i,k_1,1}}, \sum_{k_1}{\lambda_{i,k_1,2}}, \cdots, \sum_{k_1}{\lambda_{i,k_1,K}}\right]^T, 
\end{align}
\normalsize
and $h_{i,v}^T\lambda_{v}$ represents an approximation to the expectation under the $q$ distribution of the normalising function of the softmax distribution for a node $v$.  This $h_{i,v}$ is given by:
\small
\begin{equation}
	h_{i,v} = \sum_c{\exp\left(\frac{\eta_c}{n_v}\right)\prod_{\substack{j\neq i:\\ v\in \left\{s_j,r_j\right\}}}{\left(\sum_k{\lambda_{v,k}\exp\left(\frac{\eta_{c,k}}{n_v}\right)}\right)}}.
\end{equation}
\normalsize

$\eta$ is found using conjugate gradient to optimise the free energy terms of \eqref{freeE} corresponding to $\eta$: 
\small
\begin{equation}
	\mathcal{F}_{[\eta_{1:C}]} = \sum_{v}{\eta_{y_{v}}^T\bar{\lambda}_v - \log \sum_c{\prod_{\substack{i:v\in\\ \{s_i,r_i\}}}{\left(\sum_k{\lambda_{v,k}\exp\left(\frac{\eta_{c,k}}{n_v}\right)}\right)}}},
\end{equation}
\normalsize
where $\bar{\lambda}_v = \frac{1}{n_v}\sum_i{\lambda_{s_i}\delta_{s_i,v} + \lambda_{r_i}\delta_{r_i,v}}$.  Conjugate gradient requires the following derivatives:
\small
\begin{align}
  \frac{\partial \mathcal{F}_{[\eta_{1:C}]}}{\partial \eta_{c,k}} =& \sum_v{\bar{\lambda}_v\delta_{y_{v},c}} \notag\\
  & - \sum_v{\frac{\prod_{i:v\in\{s_i,r_i\}}{\left(\sum_l{\lambda_{v,l}\exp\left(\frac{\eta_{c,l}}{n_v}\right)}\right)}}{\sum_c{\prod_{i:v\in\{s_i,r_i\}}{\left(\sum_l{\lambda_{v,l}\exp\left(\frac{\eta_{c,l}}{n_v}\right)}\right)}}}}\notag\\
  & \times \sum_{i:v\in\{s_i,r_i\}}{\frac{\frac{1}{n_v}\lambda_{v,k}\exp\left(\frac{\eta_{c,k}}{n_v}\right)}{\sum_l{\lambda_{v,l}\exp\left(\frac{\eta_{c,l}}{n_v}\right)}}}.
\label{eq:conjgrad}
\end{align}
\normalsize

Predicting unlabeled nodes requires inference of the network positions given the rest of the network.  As the class label is unknown the inference is performed as above but without the terms involving $\eta$.  Classification of a test node is given by:
\small
\begin{equation}
	y_v^* = \arg \max_{y \in \{1,...,C\}} E_q \left[ \eta_y^T \bar{z_v} \right] = \arg \max_{y \in \{1,...,C\}} \eta_y^T \bar{\lambda}_v.
\end{equation}
\normalsize

\section{Experiments}
\subsection{Data}
Networks generated from four real word datasets were examined in this work: a citation network, a feeding web network and two word networks.  All of the networks are directed.  Each of the datasets have a different underlying structure with respect to the given classification task.  The first network is the Cora citation network \cite{sen:aimag08}, a popular dataset for collective classification comprising of 2708 nodes representing scientific papers and 5429 links representing the citations between them.  The classification task is to assign each paper one of 7 subject categories. 

The second network is a word network made up of the 112 most frequently occurring adjectives and nouns in Charles Dickens' novel \textit{David Copperfield} \cite{Newman:words}.  The words are linked if they appear adjacent to each other in the text.

The third network is also a word network, this time the Brown corpus\footnote{Available from http://nltk.googlecode.com/svn/trunk/nltk\_data/index.xml}\cite{francis79browncorpus}, which is a tagged corpus of present-day edited American English across various categories.  In this work a network was created using words from the News category which occurred at least 10 times and were tagged as either verb, adverb, pronoun, noun or adjective.  This resulted in a network of 990 words with 6157 links between them.  

The forth network is a food web of 463 species in the Weddell Sea in the Antarctic where the 1939 edges point to each predator from its prey\footnote{The Weddell Sea food web is available to download as part of larger dataset at http://www.esapubs.org/archive/ecol/E086/135/default.htm} \cite{foodweb}. The variable used for classification is the feeding type which takes 6 values, namely primary producer, omnivorous,
herbivorous/detrivorous, carnivorous, detrivorous, and carnivorous/detrivorous.


\subsection{Classification Performance} 
Experiments were run to investigate the classification performance between the models and as the maximum number of roles, $(K)$, was varied.  Note that for the SBM the value of $K$ was required to be fixed as equal to the number of classes $(C)$.  In all experiments 50\% of the nodes were used for training.  

Performance is measured according to the macro-averaged F1 measure given by:
\begin{displaymath}
	F_1 = \frac{2TP}{2TP + FN + FP},
\end{displaymath}
where TP, FN, and FP correspond to the true positive, false negative and false positive rates respectively.  The F1 measure represents the harmonic mean of the precision and recall values.  For the multi-class problems the macro-average is used - i.e. the F1 score is calculated for each class and then averaged.  This removes the bias in accuracy due to different class sizes in the datasets.  Each experiment was run 100 times and the performance scores reported reflect the average over these runs.

\begin{figure}%
\adjincludegraphics[width=.49\columnwidth,trim=37 0 50 0,clip=true]{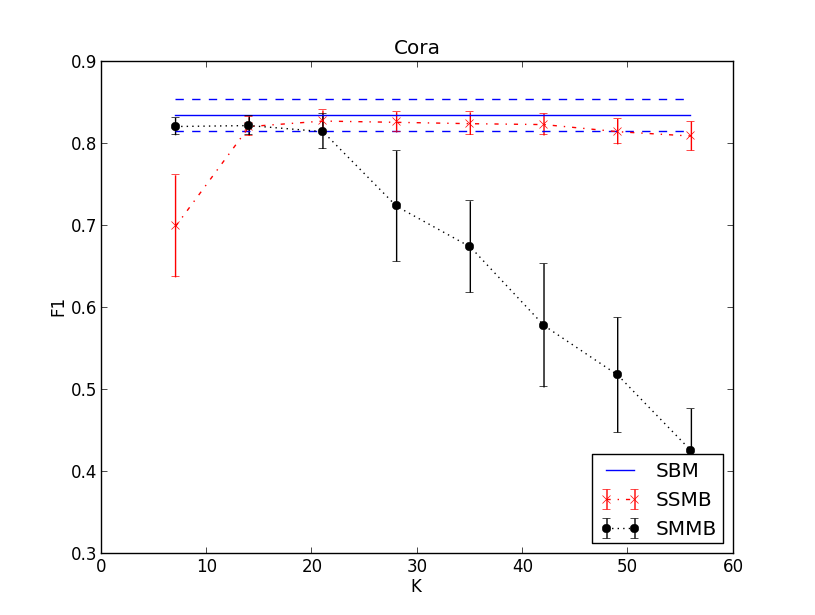}%
\adjincludegraphics[width=.49\columnwidth,trim=37 0 50 0,clip=true]{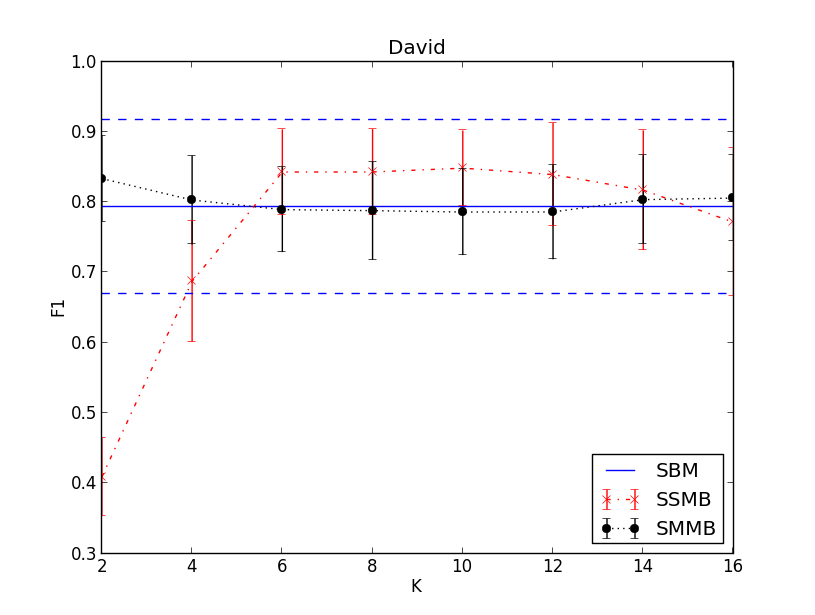}\\%
\adjincludegraphics[width=.49\columnwidth,trim=37 0 50 0,clip=true]{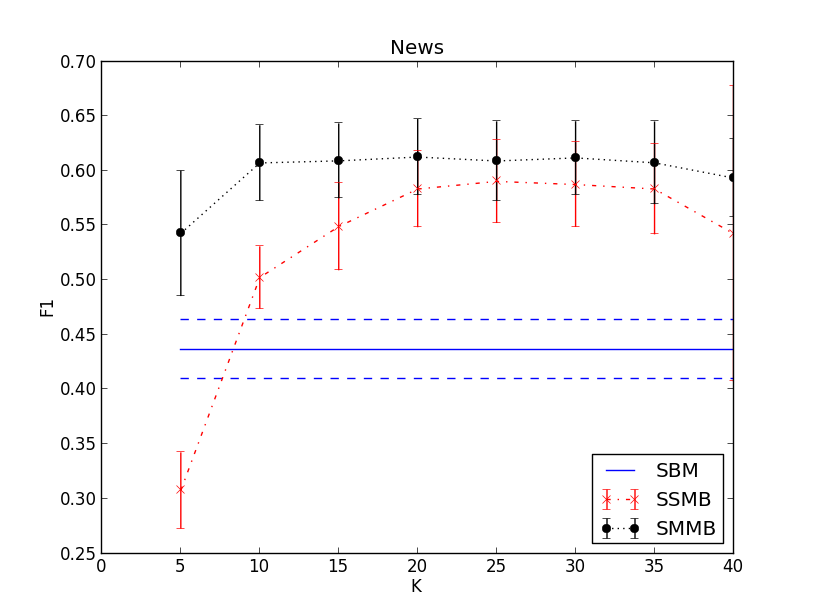}%
\adjincludegraphics[width=.49\columnwidth,trim=37 0 50 0,clip=true]{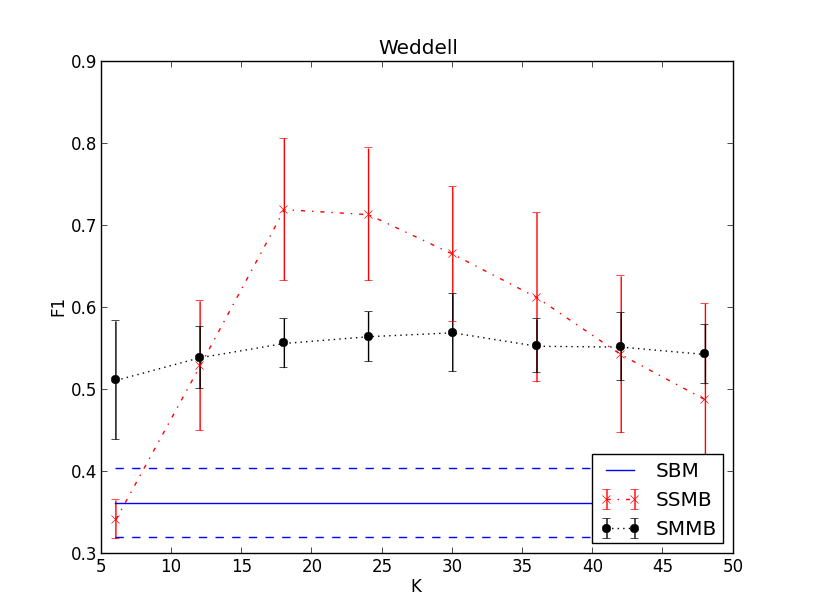}%
\caption{Classification performance of the 3 supervised blockmodel approaches as the parameter K (maximum roles) is varied.}%
\label{classperf}%
\end{figure}

Figure \ref{classperf} shows the performance of the three models across the different networks.  For the \textit{Cora} and \textit{David} networks it can be seen that the SBM model performs well and that there is no real advantage to using one of the other models.  The \textit{Cora} network is highly assortative and the \textit{David} network is highly disassortative, however, even though the networks are very different in structure they both contain homogeneous classes and so each class can be modelled with a single network role.  

The other two networks, News and Weddell, showed that the supervised (single and mixed membership) models offered a significant improvement over the SBM.  This suggests that there exists some heterogeneity in the interaction patterns within classes.  For the News dataset, the mixed membership model performs a little better than the single membership. On the other hand, in the Weddell dataset the single membership model performs a lot better than the mixed membership, however the single membership model has a much higher variance in performance compared to the more stable but less accurate mixed membership model.  

\begin{figure}%
\centering
\adjincludegraphics[width=.49\columnwidth,trim=37 0 50 0,clip=true]{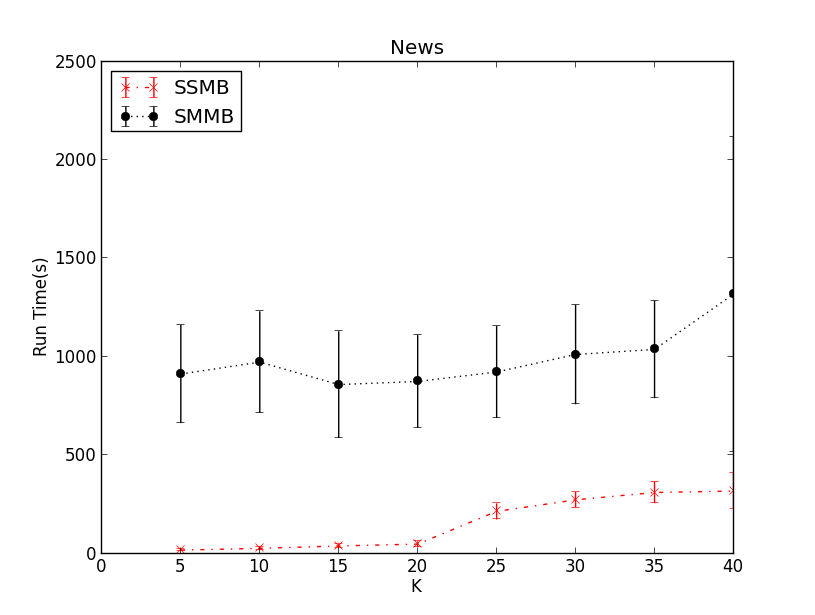}%
\adjincludegraphics[width=.49\columnwidth,trim=37 0 50 0,clip=true]{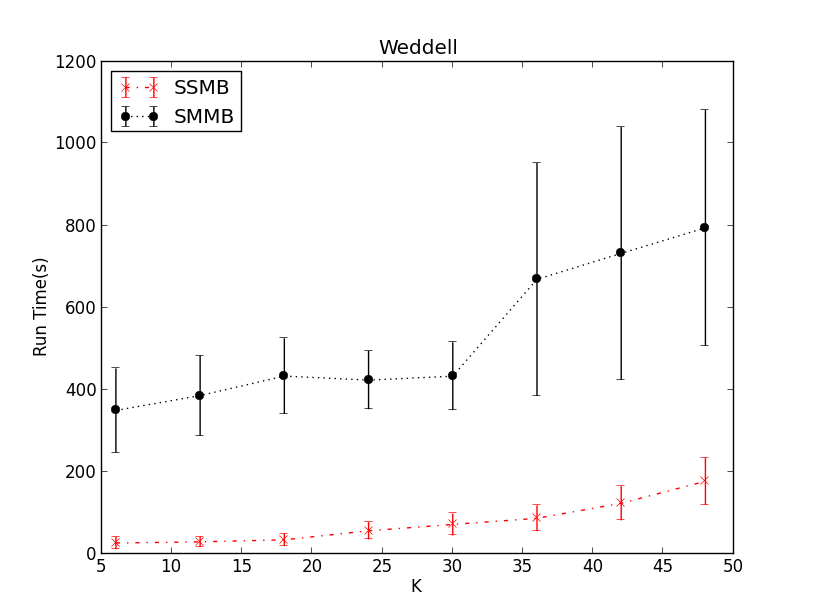}%
\caption{Average run times in seconds for the two supervised blockmodels as the parameter K (maximum roles) is varied.}%
\label{runtimes}%
\end{figure}

Figure \ref{runtimes} shows the average run times of the supervised blockmodels which is a factor of the computational complexity of the inference updates and the algorithm convergence times.  It can be seen that the mixed membership (SMMB) model is has a significantly longer run time that the single membership model due to the computational cost of finding the Softmax parameters $\eta$ in the conjugate gradient step in \eqref{eq:conjgrad}.


\subsection{Summary Networks}
\begin{figure}%
\begin{minipage}[t]{0.43\linewidth}
\includegraphics[width=\textwidth]{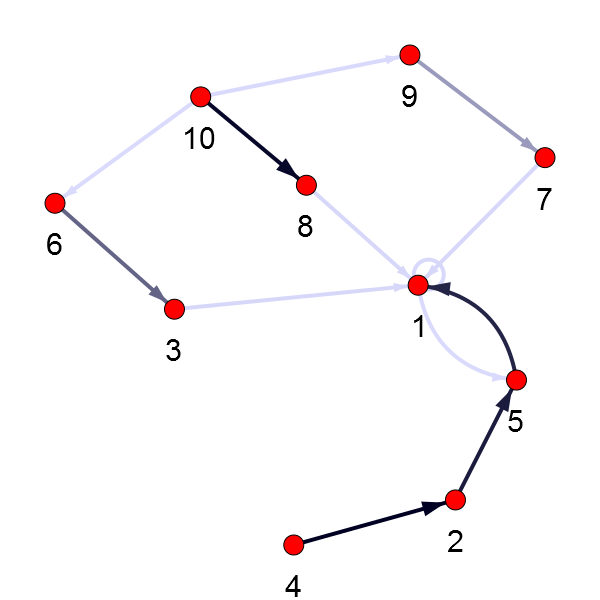}%
\caption{Summary network of the discovered role interactions in the word network created from News articles in the Brown Corpus.  Darker arrows indicate higher frequency of interaction.}%
\label{brownsumnet}%
\end{minipage}
\hspace{0.25cm}
\begin{minipage}[t]{0.53\linewidth}
\adjincludegraphics[width=\textwidth,trim=0 20 90 0,clip=true]{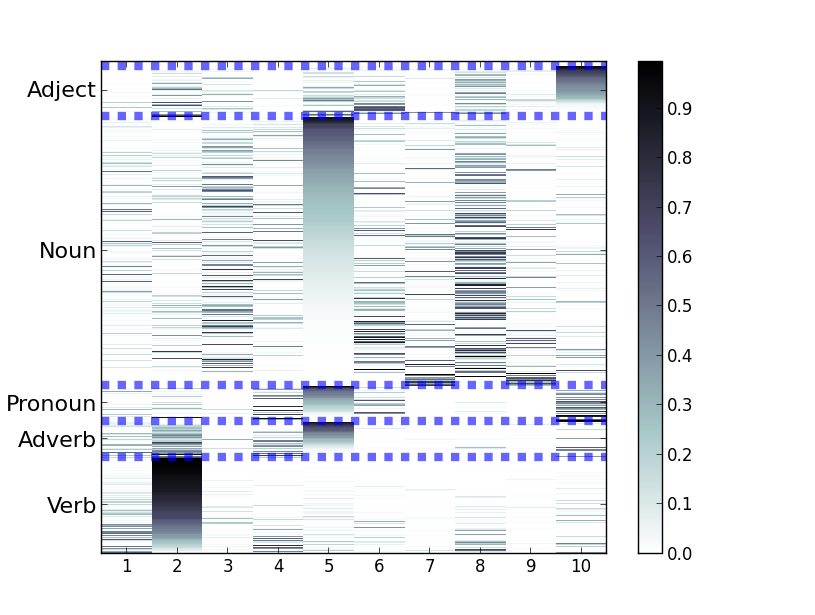}%
\caption{Node role assignment matrix for the News dataset.  Each row represents a node and its posterior distribution over roles.  Rows are ordered by class to highlight correspondence between class and position.}%
\label{brownclassrole}%
\end{minipage}
\end{figure}

\begin{figure}%
\centering
\adjincludegraphics[width=.245\columnwidth,trim=80 30 75 50,clip=true]{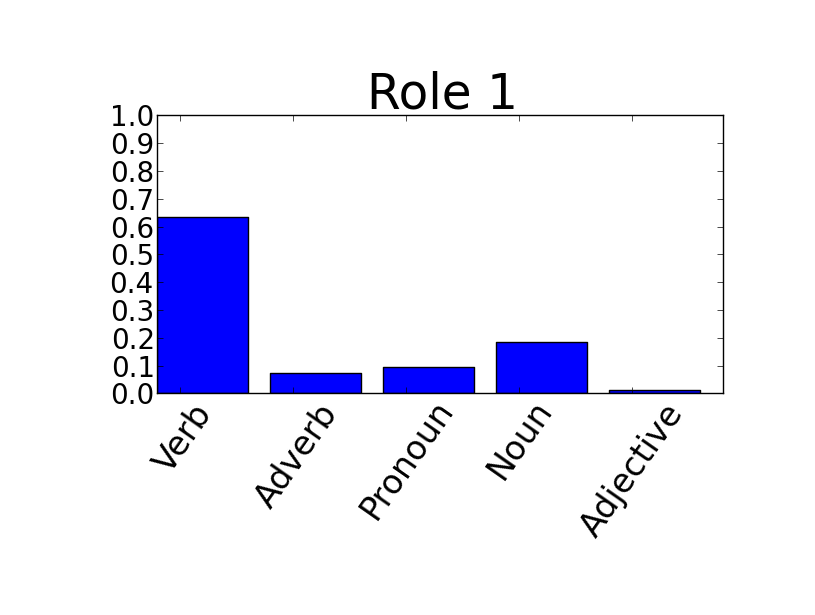}%
\adjincludegraphics[width=.245\columnwidth,trim=80 30 75 50,clip=true]{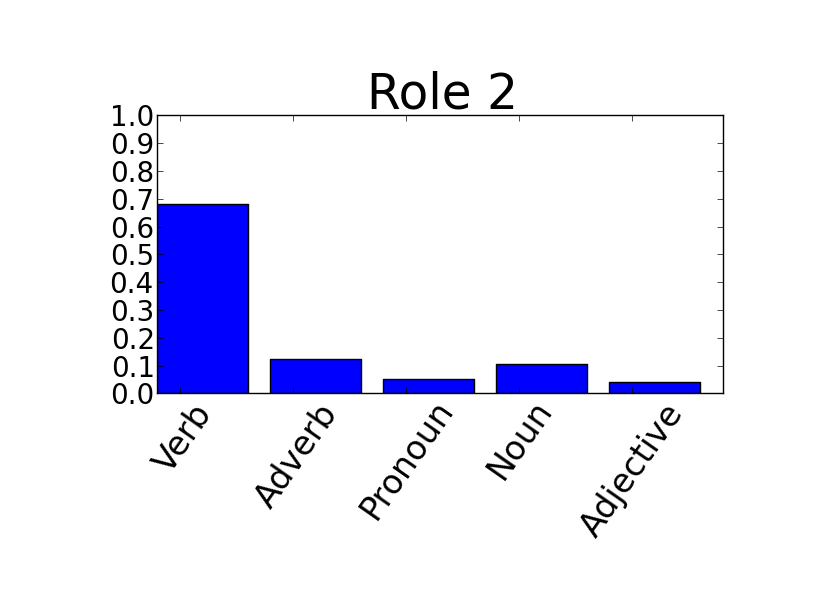}
\adjincludegraphics[width=.245\columnwidth,trim=80 30 75 50,clip=true]{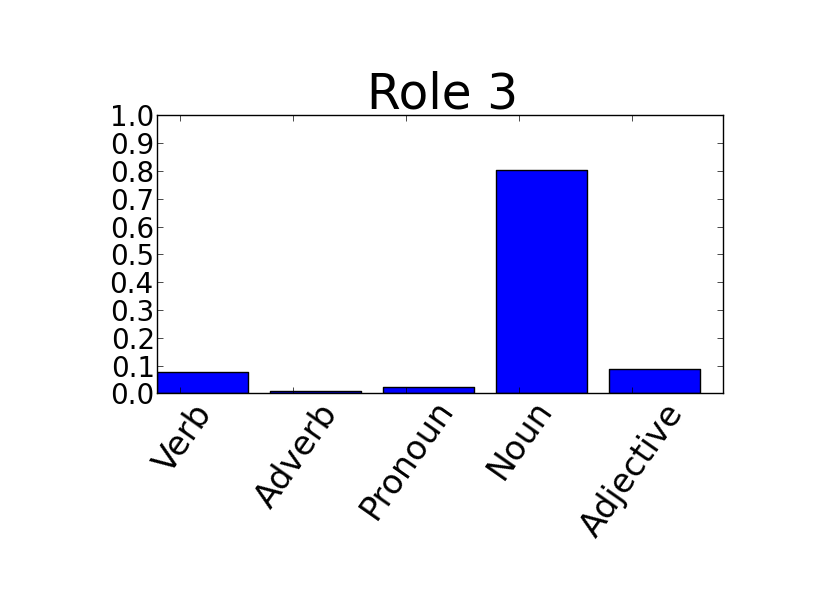}
\adjincludegraphics[width=.245\columnwidth,trim=80 30 75 50,clip=true]{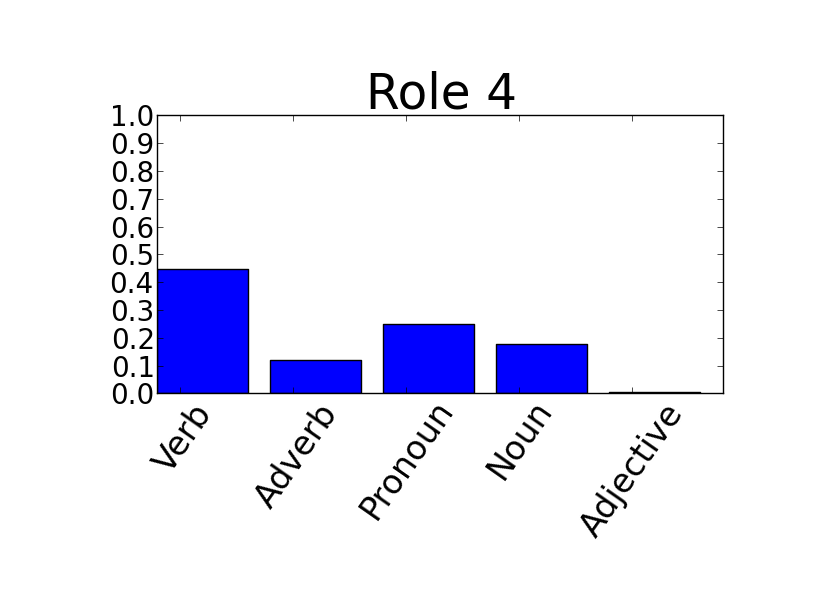}\\
\adjincludegraphics[width=.245\columnwidth,trim=80 30 75 50,clip=true]{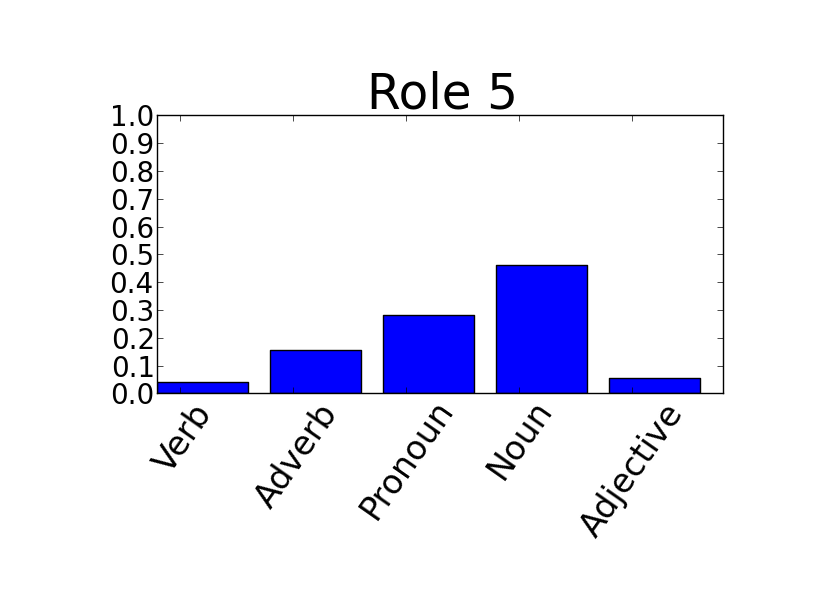}
\adjincludegraphics[width=.245\columnwidth,trim=80 30 75 50,clip=true]{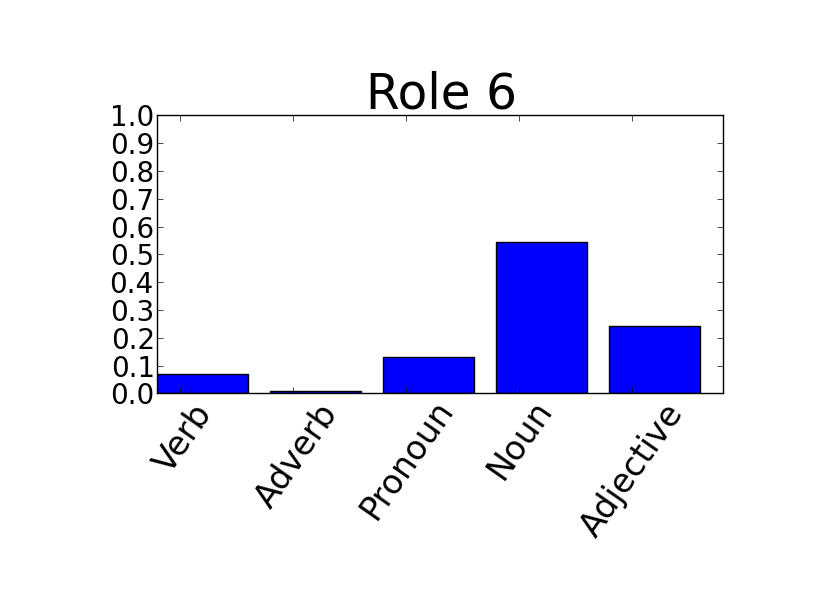}%
\adjincludegraphics[width=.245\columnwidth,trim=80 30 75 50,clip=true]{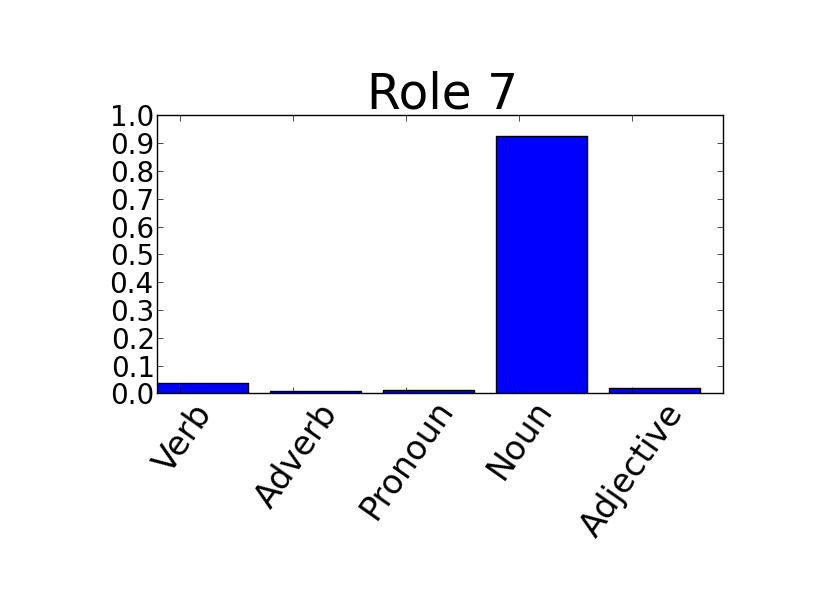}
\adjincludegraphics[width=.245\columnwidth,trim=80 30 75 50,clip=true]{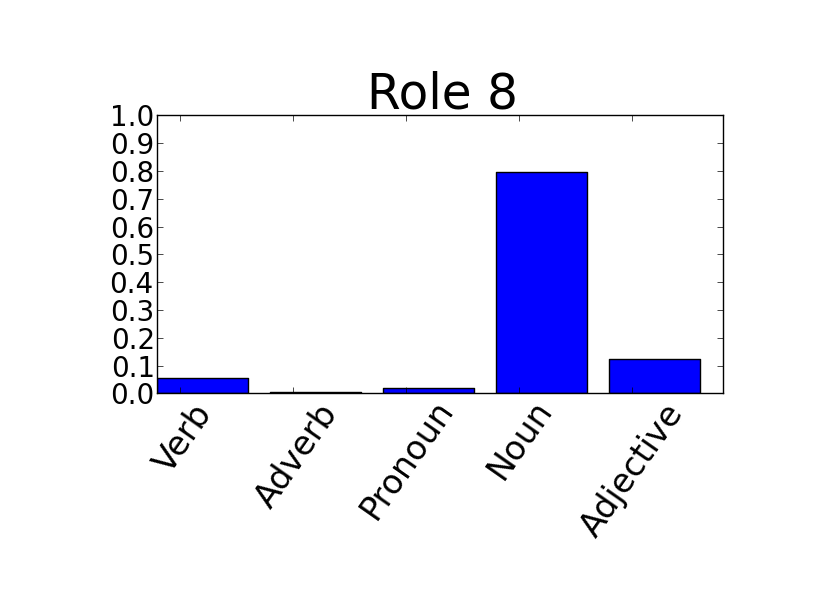}\\
\adjincludegraphics[width=.245\columnwidth,trim=80 30 75 50,clip=true]{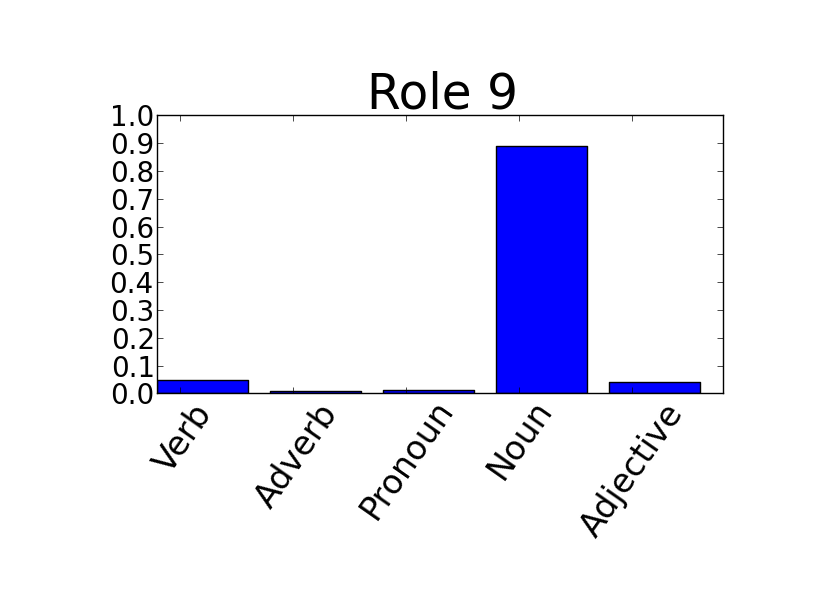}
\adjincludegraphics[width=.245\columnwidth,trim=80 30 75 50,clip=true]{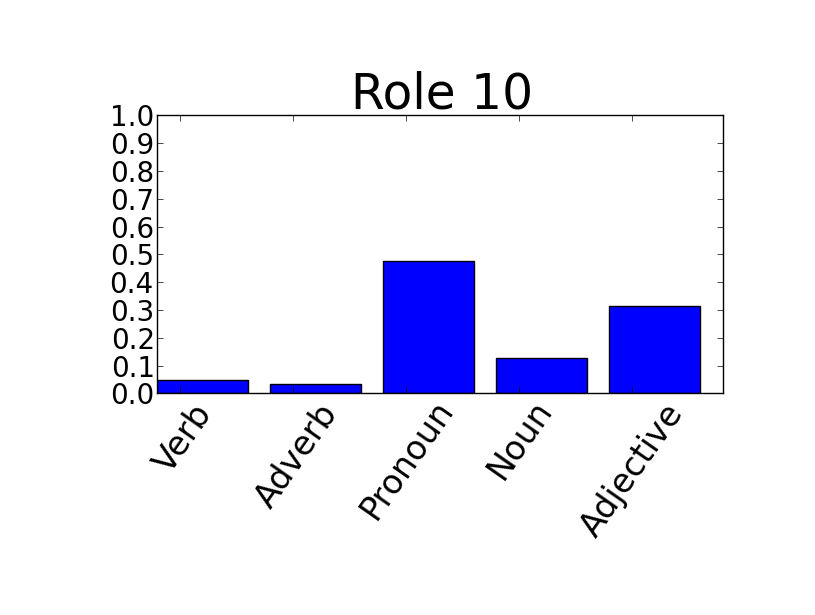}
\caption{Distributions over classes for each role in the News word network from the Brown corpus.}%
\label{roledistclass}%
\end{figure}

This section describes how the learned blockmodel can be used to understand the structure of the network with respect to the classes.  The analysis will be on the mixed membership model although a similar analysis can be conducted with the SSMB.    Focusing on the News network created from the Brown corpus, Figure \ref{brownsumnet} shows the summary network of how the identified network roles interact.  The colour of the lines indicate the probability of observing a link type, where darker edges represent more likely interactions.  Figure \ref{brownclassrole} shows a visualisation of the distribution over roles (columns) for each node (rows) in the News network.  By ordering the nodes by class it is possible to get an overall picture of the relationship of classes and network roles.  Figure \ref{roledistclass} shows the distribution over classes for each of the 10 network roles.  Using this information together it is possible to identify patterns in the connectivity of the classes and therefore in the ordering of the classes of words in the News corpus.  For example, it can be seen that Roles 1 and 2 are usually verbs and that there is a chain of frequently co-occurring roles 4-2-5-1.  Comparing the Roles 4 and 5 it can be seen that verbs (appearing more in 4 than 5) can come before other verbs but unlikely to be between two verbs.  Pronouns and Noun can come before and between verbs and Adverbs are only associated with Verbs (they do not appear in any other network role).  Other relationships that can be seen are that Nouns occur together (Roles 7 and 9) and that Pronouns and Adjectives precede Nouns (Roles 10 and 8 respectively).

%

\section{Discussion}
This work has demonstrated how the pattern of interactions alone can be used to classify unlabelled instances in relational data.  For simple cases this can be achieved with the well studied Stochastic Blockmodel (SBM) by considering the network roles as classes.  In cases where classes exhibit heterogeneity in their interactions more complex models are required.  A small modification to the Stochastic Blockmodel results in the Supervised Single Membership Blockmodel (SSMB) which can give significantly better classification performance.  The Supervised Mixed Membership Model (SMMB) also performs well but does so at a significantly higher computational cost.  Based on the few examples presented here it seems that the benefit of the mixed membership model is outweighed by its computational complexity, however more work is required to confirm this.  Finally, supervised blockmodels not only provide good classification performance but also an interpretable model to explore the structure of the data and the relationship between and within classes.

\bibliographystyle{splncs}
\bibliography{sigproc} 

\end{document}